\documentclass[10pt,twocolumn,letterpaper]{article}

\usepackage{iccv}
\usepackage{times}
\usepackage{epsfig}
\usepackage{graphicx}
\usepackage{amsmath}
\usepackage{amssymb}
\usepackage{multirow}
\usepackage{colortbl}
\usepackage{tabulary}
\usepackage{wrapfig}
\usepackage{etoolbox}
\usepackage{pifont}
\usepackage{makecell}
\usepackage{floatrow}
\usepackage{subfig}
\usepackage{booktabs}
\usepackage{rotating}

\usepackage{tablefootnote}
\definecolor{LightCyan}{rgb}{0.8,1,1}
\definecolor{LightGreen}{rgb}{0.6,0.9,0.7}
\definecolor{royalblue}{rgb}{0.0, 0.14, 0.4}
\usepackage{xcolor, soul}
\sethlcolor{LightGreen}
\colorlet{LightCyan}{LightCyan}
\usepackage{colortbl}
\usepackage{tabularx}
\usepackage{mathtools}
\usepackage{ragged2e}
\usepackage[export]{adjustbox}
\usepackage{wrapfig}
\usepackage{caption}
\usepackage{floatrow}
\def\boxit#1{%
  \smash{\fboxsep=0pt\llap{\rlap{\fbox{\strut\makebox[#1]{}}}~}}\ignorespaces
}

\usepackage[breaklinks=true,bookmarks=false]{hyperref}

\iccvfinalcopy 

\def\iccvPaperID{****} 

\ificcvfinal\fi

\begin{document}

\title{\textsc{AD-CLIP}: \underline{A}dapting \underline{D}omains in Prompt Space Using \underline{CLIP}}

\author{Mainak Singha$\thanks{equal contribution}$ \and Harsh Pal$^{*}$ \and Ankit Jha$^{*}$ \and Biplab Banerjee$^{}$
\and
Indian Institute of Technology Bombay, India
\and
{\tt\small \{mainaksingha.iitb, palharsh.india, ankitjha16, getbiplab\}@gmail.com}}
\maketitle
\ificcvfinal\thispagestyle{empty}\fi

\begin{abstract}
Although deep learning models have shown impressive performance on supervised learning tasks, they often struggle to generalize well when the training (source) and test (target) domains differ. Unsupervised domain adaptation (DA) has emerged as a popular solution to this problem.
However, current DA techniques rely on visual backbones, which may lack semantic richness. Despite the potential of large-scale vision-language foundation models like CLIP, their effectiveness for DA has yet to be fully explored.
To address this gap, we introduce \textsc{AD-CLIP}, a domain-agnostic prompt learning strategy for CLIP that aims to solve the DA problem in the prompt space. We leverage the frozen vision backbone of CLIP to extract both image style (domain) and content information, which we apply to learn prompt tokens.
Our prompts are designed to be domain-invariant and class-generalizable, by conditioning prompt learning on image style and content features simultaneously. We use standard supervised contrastive learning in the source domain, while proposing an entropy minimization strategy to align domains in the embedding space given the target domain data.
We also consider a scenario where only target domain samples are available during testing, without any source domain data, and propose a cross-domain style mapping network to hallucinate domain-agnostic tokens. Our extensive experiments on three benchmark DA datasets demonstrate the effectiveness of \textsc{AD-CLIP} compared to existing literature. Code is available at \url{https://github.com/mainaksingha01/AD-CLIP}
\end{abstract}

\section{Introduction}
\label{sec:intro}
\begin{figure}[ht!]
    \centering
    \includegraphics[width=0.8\textwidth]{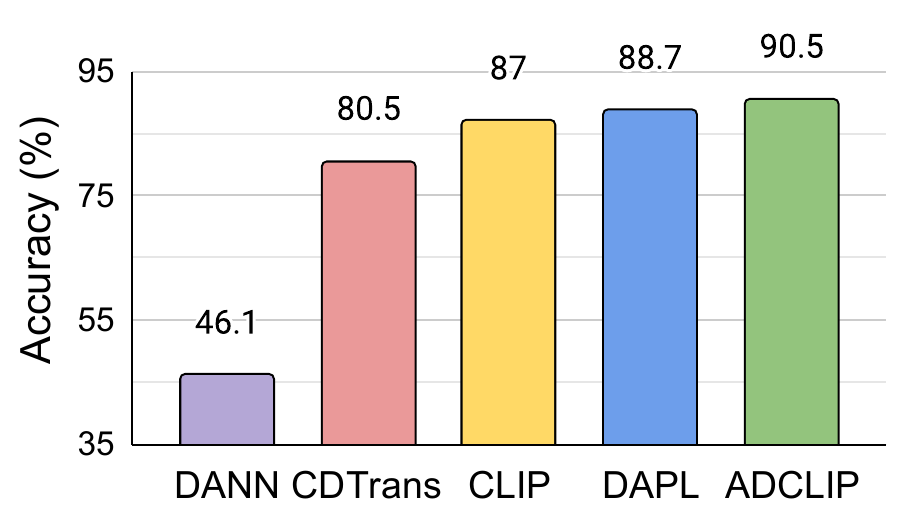}
    \vspace{-1.5mm}
    \caption{\textcolor{royalblue}{We compare the performance of \textsc{AD-CLIP} for the Office-Home \cite{officehome} dataset, with different type of UDA methods, e.g. convnets-based DANN \cite{dann}, Transformer-based CDTrans \cite{cdtrans}, pre-trained CLIP \cite{clip} without prompt learning and, DAPL \cite{dapl}, a prompt learning-based DA technique.}}
    \label{fig:teaser}
    \vspace{-3mm}
\end{figure}
The use of deep convolutional neural networks (convnets) has led to significant advancements in visual recognition tasks within supervised learning settings \cite{krizhevsky2017imagenet}. These models can learn discriminative, data-driven features from large sets of training samples. However, they are vulnerable to the domain-shift problem, which arises when training and test samples come from different distributions, causing the probable approximately correct (PAC) \cite{haussler1993probably} assumption to fail.
One potential solution to this issue is domain adaptation (DA) \cite{yao2015semi,daume2010frustratingly,baktashmotlagh2013unsupervised}, a form of transductive transfer learning. DA leverages both labeled source data and unlabeled target data to create a domain-agnostic embedding space. This space can be used to train classifiers on the source domain, which can accurately classify target samples. Numerous DA techniques are available in the literature, including adversarial, entropy minimization, and statistical distance optimization approaches \cite{wang2018deep}.
\textit{However, current models typically rely on convnets \cite{resnet} purely trained on visual data, which often lack semantic richness, thus producing sub-optimal performance in critical situations} (see Fig. \ref{fig:teaser}).

Large-scale vision-language models \cite{clip,align} have emerged as the \textit{de-facto} feature extractor in computer vision nowadays. These models are trained on vast quantities of image-text pairs, where class labels are represented as textual prompts (e.g., \texttt{a photo of a [CLS]}). This results in a joint embedding space with rich semantics, facilitating excellent generalization and zero-shot classification performance.
Although the CLIP \cite{clip} model has demonstrated impressive results, designing task-specific prompts can be challenging. Subsequent works \cite{coop, cocoop} have focused on learning prompts in a data-driven manner, primarily utilizing visual information from CLIP's frozen image backbone. Despite the success of these approaches, they have yet to be meaningfully applied for cross-domain inference tasks, with only a few works focusing on domain generalization \cite{stylip}.
\textit{In opposition, our focus is on solving the DA problem by leveraging the semantic richness of CLIP. To achieve this goal, we aim to use the pre-trained backbones of CLIP without fine-tuning, but propose to learn prompts that can capture the domain and class distributions well and are domain-agnostic, by introducing only a small set of learnable parameters.}


\vspace{0.2cm}
\textbf{Our proposed \textsc{AD-CLIP}:} 
In this paper, we introduce a novel framework called \textsc{AD-CLIP} to address our research questions. Our main objective is to design prompts that can generalize well across the source and the target domains. To achieve this, we propose a method to learn new prompts that consist of two types of tokens: i) \textit{Domain token}: To incorporate domain knowledge, we introduce a token that captures the style information of both domains. Style corresponds to the feature statistics obtained from CLIP's image encoder \cite{li2017demystifying}, and we propose a way to combine the multi-scale style features through a \textit{style projector}.
ii) \textit{Image-specific tokens}: To learn the visual distributions well in the semantic space and obtain a distribution of prompts per class, we leverage the visual feature responses from the different layers of CLIP's vision encoder to initiate learning of image-conditioned tokens. We note that we consider the multi-scale features in this aspect, as they can better characterize the underlying visual concepts at multiple abstractions. We introduce a set of \textit{content projectors} for this purpose, and all the projectors are trained contrastively given the prompt and image embeddings. For aligning the target domain data with the source counterparts, we propose simple yet effective entropy minimization characteristics given the similarity distributions between the image embeddings and the prompt embeddings for all the classes while aligning the cross-domain prompt embeddings through optimizing a measure of distributions divergence.

\begin{figure*}
    \centering
    \includegraphics[scale=0.9]{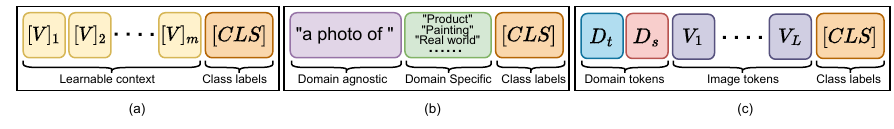}
    \caption{\textcolor{royalblue}{We highlight the differences between our prompts from the literature. a) CoOp \cite{coop} directly learns the prompt tokens from random vectors and may not be suitable for DA as it does not concern domain independence, b) Another possibility is to manually include the \texttt{domain name} into manually defined prompts, but this information may not be readily available, c) \textsc{AD-CLIP} introduces an automatic solution by leveraging the visual space to define the domain-agnostic and image-conditioned tokens.}}
    \label{fig:prompting}
\end{figure*}

During the inference stage, we often only have access to the test images from the target domain, without any access to the source domain data. This creates a challenge when trying to define the domain-driven prompt token without the knowledge of the source domain, which is only available during training. To overcome this, we propose a novel approach that involves hallucinating the source domain characteristics based on the target domain properties, using a \textit{target-to-source style prediction network}. We train this network by passing the style features of the target domain images, which are then used to generate the corresponding source style information. Our approach distinguishes itself from other prompt learning methods described in the literature \cite{coop, cocoop} in the way we propose to generate the prompts while ensuring domain independence and generalizability jointly (see Figure \ref{fig:prompting}). Our \textbf{significant contributions} are summarized as follows:

\noindent [-] We propose a solution to the challenging domain adaptation problem using prompt learning within the context of CLIP. Our primary focus is to ensure that the prompts are not biased towards a specific domain and account for the visual variations in the data.

\noindent [-] To achieve this, we propose a novel prompt learning scheme that entirely leverages the visual encoder of CLIP and introduces a small set of learnable projector networks. We also propose a new entropy minimization-based criterion for domain alignment. Furthermore, we address the scenario where source domain data are not available during inference and develop a method to approximate the prompts for the target images.

\noindent [-] Through extensive experiments on three widely-used benchmark DA datasets, namely Office-Home \cite{officehome}, VisDA \cite{visda}, and mini-DomainNet \cite{domain_net}, we demonstrate the superior performance of \textsc{AD-CLIP} over state-of-the-art alternatives.

\section{Related Works}
\label{sec:literature}
\noindent\textbf{Unsupervised Domain Adaptation:}
DA is the process of adapting a machine learning model trained on a source domain to a target domain where the data distributions may differ. The literature is rich with a plethora of DA approaches, including distribution alignment based on sub-space alignment, pseudo-labeling, or adversarial techniques, among others.
For example, Maximum Mean Discrepancy (MMD) \cite{MMD} reduces the distance between the distributions of the source and target domains in the kernel space. Another popular approach is DANN \cite{dann}, which involves adding a domain classifier to the deep neural network, enabling it to learn to distinguish between source and target domain data. CyCADA \cite{cycada} utilizes cycle-consistent adversarial learning to align the feature distributions. CDTrans \cite{cdtrans} uses cross-attention and two-way center-aware labeling in Transformers \cite{transformer} for domain alignment, making it robust to noisy label pairs. A more detailed discussion on DA can be found in \cite{wang2018deep, zhao2020review}.
Recent approaches have started considering vision-language models for solving the DA task given their enhanced feature space. \textit{However, the existing sole method in this regard, DAPL \cite{dapl}, uses ad-hoc prompting to learn disentangled domain and category representations. DAPL \cite{dapl} manually includes the domain information, which is unrealistic in some cases. Additionally, DAPL \cite{dapl} ignores the visual distributions of the classes, causing overfitting.}\\


\noindent\textbf{Vision-Language models and Prompt Learning:} 
The large-scale vision-language models (VLMs) integrate visual and textual inputs to achieve a more comprehensive understanding of the world, leading to better performance in various computer vision tasks.
They typically rely on pre-trained language models, such as BERT \cite{bert} and GPT \cite{gpt}, to encode textual inputs, while the visual inputs are processed using convnets or vision transformers. Some of the popular VLMs are CLIP \cite{clip} and VisualBERT \cite{visualbert}.

In a similar spirit, prompt learning for VLMs is a technique that has gained increasing attention in computer vision, which involves leveraging pre-trained language models to provide valuable insights for downstream tasks through prompts. 
Several recent studies have explored the use of prompt learning, such as CoOp \cite{coop}, and CoCoOp \cite{cocoop}, which use conditional prompts to improve the model's generalization capabilities. AutoPrompt \cite{autoprompt} explores tokens with the most significant gradient changes in the label likelihood to automate the prompt generation process. Whereas, APPLeNet \cite{applenet} addresses the problem of DG in remote sensing by introducing prompt learning. Another recent study, MaPLe \cite{maple}, proposes multi-modal prompt learning to avoid possible overfitting. \textit{However, none of the existing prompting techniques is tailored for the DA task except DAPL \cite{dapl}, which is majorly hand-crafted. In contrast, we propose a more robust prompt learning technique while ensuring domain independence and improving the adaptation capabilities both in image and text feature space.} 



\section{Proposed Methodology}
\noindent \textbf{Problem Definition}:
The DA problem involves a source domain with the image-label pairs, $\mathcal{D}^{\mathcal{S}_l} = \{x_i^{\mathcal{S}_l}, y_i^{\mathcal{S}_l}\}_{i=1}^{N_{\mathcal{S}_l}}$ ($x_i \in \mathcal{X}^s$, $y_i \in \mathcal{Y}$), where the labeled data follows the joint distribution $\mathcal{P}^{\mathcal{S}_l}_{data}$, and a target domain with unlabeled images, $\mathcal{D}^{\mathcal{T}_u} = \{x_j^{\mathcal{T}_u}\}_{j=1}^{N_{\mathcal{T}_u}}$, where the unlabeled data follows the distribution $\mathcal{P}^{\mathcal{T}_u}_{data}$, respectively. It is important to note that $\mathcal{P}^{\mathcal{T}_u}_{data}$ is not equal to $\mathcal{P}^{\mathcal{S}_l}_{data}$, leading to domain shift. The number of images in the source and target domains is denoted by $N_{\mathcal{S}_l}$ and $N_{\mathcal{T}_u}$, respectively. Also, in the closed-set approach that we follow, $\mathcal{S}_l$ and $\mathcal{T}_u$ share the same label space $\mathcal{Y}$. Under this setting, the objective is to learn a classifier $f: \mathcal{X}^s \rightarrow \mathcal{Y}$ that performs well on $\mathcal{T}_u$ by leveraging $\mathcal{S}_l$ and $\mathcal{T}_u$, which requires overcoming the distributional differences between $\mathcal{D}^{\mathcal{S}_l}$ and $\mathcal{D}^{\mathcal{T}_u}$.\\

\begin{figure*}[t]
    \centering
    \includegraphics[scale=0.8]{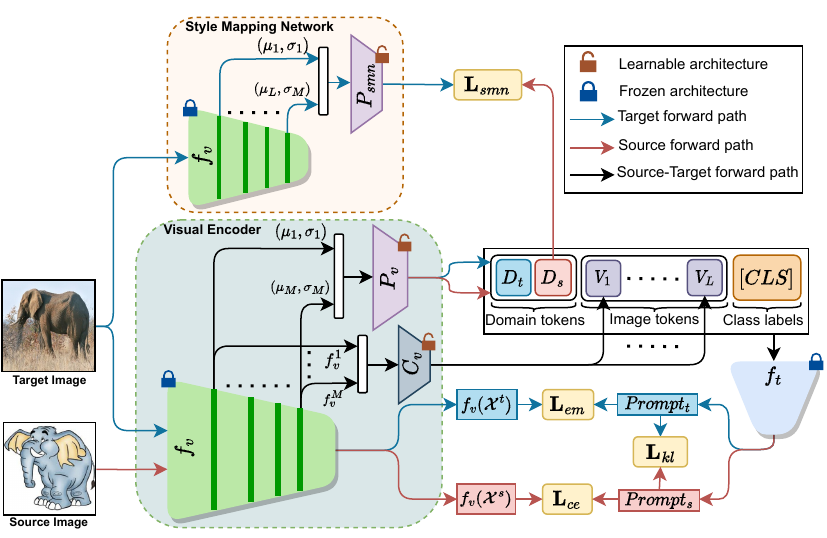}
    \caption{ \textcolor{royalblue}{\textbf{The architecture of AD-CLIP} is based on the frozen CLIP backbones $f_v$ and $f_t$. For prompt token learning, we introduce the new vision and text projectors $P_v$ and $C_v$, respectively, which encoder the style and content information from the different layers of $f_v$. The style mapping network, $P_{smn}$, approximates the source domain style information from the target domain features. Best viewed in color.}}
    \label{fig:AD-CLIP}
\end{figure*}

\noindent\textbf{Overview of \textsc{AD-CLIP}}:
In the following, we delve into the details of \textsc{AD-CLIP}. Our primary goal is to learn domain- and class-agnostic prompts that lead to a discriminative and domain-aligned semantic embedding space. To achieve this, we utilize the frozen vision and text backbones of CLIP, referred to as $f_v$ and $f_t$, respectively, both of which rely on transformers. To enable the learning of prompt tokens using visual information from different layers of $f_v$, we introduce learnable style and content projectors, $P_v$ and $C_v$, respectively. Specifically, given $f_v$ comprising $M$ encoder layers, $P_v$ and $C_v$ facilitate prompt learning in parallel by separately looking into the image domain and content properties. Furthermore, we incorporate the target-to-source style mapping unit $P_{smn}$ to hallucinate source style features from the target domain samples during inference. While $P_{smn}$ and $P_v$ take the form of an encoder-decoder, $C_v$ is designed to consist of a single encoder and $L$ decoders, one per prompt token, where $L$ is the context length for the prompts.

We proceed to discuss the following: i) prompt learning in \textsc{AD-CLIP} using disentangled visual style and content information, ii) the target-to-source style mapping network, and iii) the loss functions for classification and domain alignment.

\vspace{0.2cm}
\noindent\textbf{- Our proposed prompt learning}: Our objective is to learn prompts directly from the visual domain to effectively encode the visual distribution, as opposed to the static prompting technique \cite{coop}. In this regard, we have two primary objectives for addressing the DA task: i) incorporating a domain-agnostic token into the prompt to prevent domain bias, and ii) enhancing the learning of visual concepts in prompt tokens by utilizing feature responses from multiple layers of the CLIP vision encoder.
We introduce a domain-agnostic token of the form [$D_s;D_t$]. To obtain $D_s$, we pass the multi-scale style information through the shared style projector $P_v$. Precisely, the style information is represented by the first and second-order batch-wise feature statistics: $[\mu, \sigma]$. In our case, we calculate and combine $[\mu_1; \sigma_1; \cdots ; \mu_M; \sigma_M]$ from the $M$ layers of $f_v$ for a given $x$ to obtain the style vector $\bar{\mathcal{F}}(x)$.
Similarly, we define the multi-scale image content features as $\hat{\mathcal{F}}(x) = [\hat{f}_v^1(x); \cdots ;\hat{f}_v^M(x)]$ where $f_v^m$ denotes the responses from the $m^{th}$ layer. For a given batch, we consider the samples from $\mathcal{S}_l$ to produce $\bar{\mathcal{F}}_s$ and samples from $\mathcal{T}_u$ to produce $\bar{\mathcal{F}}_t$, respectively, for each $x$. $P_v$ subsequently maps $\bar{\mathcal{F}}_s$ onto $D_s$ and $\bar{\mathcal{F}}_t$ onto $D_t$.
On the other hand, $\hat{\mathcal{F}}(x)$ is passed through $C_v$ to produce $L$ image-specific context tokens $\{V_l\}_{l=1}^L$. Finally, we denote the prompt for a class $y$ with the class embedding $CLS_y$ given an image $x$ as,
\begin{equation}
    \centering
    \text{Prompt}_y(x) = \biggr[[D_t;D_s]; V_1; \cdots; V_L; [CLS_y]\biggr]
\end{equation}
\noindent \textbf{- The target to source style mapping network:} Although the model is trained in a transductive manner, where samples from both domains are used during training, it may encounter target domain samples separately during the inference stage. This absence of source domain samples during inference can hinder the generation of prompts, as the prompt relies on the domain-agnostic token. To address this challenge, we introduce a cross-domain style mapping network, denoted as $f_{smn}$. This network takes $\bar{\mathcal{F}}_t^b$ as input and learns to produce the corresponding $D_s$ given the samples for the $b^{th}$ batch. We train $f_{smn}$ using the $\ell_2$ loss, defined as follows:

\begin{equation}
    \centering
    \mathbf{L}_{smn} = \underset{f_{smn}, P_v} {\arg\min} \underset{P_{data}^{\mathcal{S}_l}, P_{data}^{\mathcal{T}_u}}{\mathbb{E}} ||D_s - f_{smn}(\bar{\mathcal{F}}_t)||_2^2
\end{equation}

\noindent \textbf{- Loss functions pertaining to domain alignment:}
We consider two loss objectives for learning the classifier for $\mathcal{S}_l$, while aligning the prompts for $\mathcal{T}_u$ with those of the source domain.
The source domain supervised contrastive loss between $f_v(x)$ and $\text{Prompt}_y$ given ($x,y$) is used for image-text mapping and is optimized through the cross-entropy loss $\mathbf{L}_{ce}$. In this regard, the prediction probability of $x$ for label $y$ is defined as,
\begin{equation}
p(y|x) = \frac{ \exp(\text{sim}(f_v(x), f_t(\text{Prompt}_y(x)))/\tau)}{\sum_{k=1}^{|\mathcal{Y}|}\ \exp(\text{sim}(f_v(x), f_t(\text{Prompt}_{y_k}(x)))/\tau)}
\end{equation}

where, `$\text{sim}$' denotes the \texttt{cosine} similarity, and $\tau$ is the temperature hyper-parameter.

Ideally, we can treat the prompts as class prototypes and aim to map the visual features onto these prototypes. This means we aim to ensure that target domain samples are aligned to one prototype while being pushed away from others to achieve domain alignment. Additionally, we want to increase the correlation between the prompts generated from the two domains, since we seek to generate a prompt-aligned semantic space.
To accomplish both tasks, we have introduced a new loss objective $\mathbf{L}_{Align}$. Our approach minimises the distribution divergence between the source and target prompt embeddings using a Kullback-Leibler (KL) divergence loss ($\mathbf{L}_{KL}$). We also constrain the similarity distribution between the visual features of the target samples and prompt embeddings to have low entropy through $\mathbf{L}_{em}$. Together, they enforce the model to produce similar types of prompt embedding distributions for both domains, while aligning each target sample to a single prompt in a discriminative fashion. '$\text{Prompt}_s$' denotes all the prompts in the source domain and likewise for $t$.

\begin{equation}
    \centering
    \begin{aligned} 
    \mathbf{L}_{Align} = &\underset{P_{v}, C_v} {\arg\min} \underset{(x,y) \in P_{data}^{\mathcal{S}_l}}{\mathbb{E}} \mathbf{L}_{em}([p(y_1|x);\cdots;p(y_{|\mathcal{Y}|}|x)]) \\ &
  \hspace{2.5cm}  + \mathbf{L}_{KL} (\text{Prompt}_t|\text{Prompt}_s)
    \end{aligned}
\end{equation}

\vspace{0.3cm}
\noindent \textbf{- Total loss:} We train \textsc{AD-CLIP} with respect all the losses mentioned above as: $\mathbf{L}_{total} = [\mathbf{L}_{ce} + \mathbf{L}_{smn} + \mathbf{L}_{Align}]$. 

Inference involves comparing the embeddings of the target samples to all the class prompt embeddings and selecting the class maximizing $p(y|x)$.

\section{Experimental Evaluations}
\noindent \textbf{Datasets descriptions:}\label{sec:dataset} We validate our model on three publicly available DA datasets.
i) \textbf{Office-Home} \cite{officehome}: This dataset is comprised of 15,500 high-quality images from four distinct domains: Art (Ar), Clip Art (Cl), Product (Pr), and Real World (Rw). Each domain contains a diverse range of objects from 65 different categories, set within both office and home environments.
ii) \textbf{VisDA-2017} \cite{visda}:
The VisDA-2017 dataset presents a more challenging scenario for synthetic-to-real domain adaptation, featuring 12 categories with 152,397 synthetic images generated by rendering 3D models from different angles and light conditions, and 55,388 real-world images collected from MSCOCO. To maintain consistency with established protocols \cite{conditional_DA,maximum_discrepency}, we use the synthetic images as the source domain and the real-world images as the target domain.
iii) \textbf{Mini-domainNet}:
Lastly, we consider a subset of the comprehensive DomainNet dataset \cite{domain_net} called Mini-DomainNet. This subset features four domains, including Clipart (c), Painting (p), Real (r), and Sketch (s), each with images from 126 categories.

\vspace{0.2cm}
\noindent \textbf{Architecture Details:}
For our experiments, we utilize three pre-trained vision encoders as $f_v$: ResNet-50 (RN50) \cite{resnet}, ViT-L/14, and ViT-B/16 \cite{vit} for validation. Meanwhile, we employ a transformer-based text encoder as $f_t$.
To facilitate our projective transformation, we implement the $P_v$ and $P_{smn}$ projector networks using a single encoder and decoder layer. On the other hand, $C_v$ consists of a dense encoder and $L$ dense decoder layers, respectively. 

\vspace{0.2cm}
\noindent \textbf{Training and evaluation protocols:} We optimize $\mathbf{L}_{total}$ using the \textit{Adam} \cite{kingma2014adam} optimizer, given a mini-batch size of $16$, and an initial learning rate of $0.01$, respectively. Finally, we report the target domain top-1 accuracy (\texttt{mean $\pm$ std.}) over three runs as the evaluation metric. We compare \textsc{AD-CLIP} against traditional DA-techniques based on vision backbones like ResNet-50 \cite{dann, gsda, spl, srdc}, pre-trained CLIP and DAPL \cite{dapl} visual features based on ViT-B/16 \cite{vit} and ViT-L/14 \cite{vit}, to name a few.

\subsection{Comparisons to the state-of-the-art}

In this section, we present the results of our extensive evaluation of \textsc{AD-CLIP} alongside several state-of-the-art methods for domain adaptation (DA) on three benchmark datasets: Office-Home, VisDA-2017, and Mini-DomainNet. We also compare \textsc{AD-CLIP} with traditional CNN-based and Transformer-based unsupervised domain adaptation (UDA) methods, as well as vision-language foundation models. The evaluation results, presented in Tables \ref{tab:officehome}-\ref{tab:minidomainnet}, demonstrate that \textsc{AD-CLIP} achieved substantial improvements on all three benchmark datasets. Specifically, it surpassed the prior best by $1.8\%$ in Office-Home \cite{officehome}, by $2.2\%$ in VisDA-2017 \cite{visda}, and by $1.6\%$ in Mini-DomainNet \cite{domain_net}, thereby establishing a new performance benchmark for domain adaptation tasks. In comparison with traditional CNN-based and Transformer-based UDA methods, along with vision-language foundation models, \textsc{AD-CLIP} consistently exhibited superior performance. Notably, \textsc{AD-CLIP} outperformed these methods across various evaluation metrics, reaffirming its effectiveness as a robust solution for domain adaptation challenges.

\begin{table*}[!ht]
    \centering
    \caption{\textcolor{royalblue}{Comparison of \textsc{AD-CLIP} with state-of-the-art methods for UDA task on Office-Home \cite{officehome} dataset. We show our results with three different vision backbones; ResNet50\cite{resnet}, ViT-B/16 \cite{vit} and ViT-L/14 \cite{vit}. Whereas, CDTrans* has used DeiT-base backbone only. The overall best accuracy and best within per backbone are indicated in bold and box respectively.}}
    \scalebox{0.6}{
    \begin{tabular}{lccccccccccccccc}
    \toprule
        
       \rowcolor{gray!20} {\textbf{Method}} & $f_v$ & {Ar$\rightarrow$Cl} & {Ar$\rightarrow$Pr} & {Ar$\rightarrow$Rw} & {Cl$\rightarrow$Ar} & {Cl$\rightarrow$Pr} & {Cl$\rightarrow$Rw} & {Pr$\rightarrow$Ar} & {Pr$\rightarrow$Cl} & {Pr$\rightarrow$Rw} & {Rw$\rightarrow$Ar} & {Rw$\rightarrow$Cl} & {Rw$\rightarrow$Pr} & {\textbf{Avg}} \\ 
        \midrule

        \cellcolor[gray]{0.9}RN-50 \cite{resnet} & & 34.9& 50.0& 58.0& 37.4& 41.9& 46.2& 38.5& 31.2& 60.4& 53.9& 41.2& 59.9& $\cellcolor[gray]{0.9}46.1$  \\

        \cellcolor[gray]{0.9}DANN \cite{dann}& & 45.6& 59.3& 70.1& 47.0& 58.5& 60.9& 46.1& 43.7& 68.5& 63.2& 51.8& 76.8& $\cellcolor[gray]{0.9}57.6$ \\

        \cellcolor[gray]{0.9}GSDA \cite{gsda}& & 61.3& 76.1& 79.4& 65.4& 73.3& 74.3& 65.0& 53.2& 80.0& 72.2& 60.6& 83.1& $\cellcolor[gray]{0.9}70.3$ \\

        \cellcolor[gray]{0.9}GVB-GD \cite{gvbgd}& & \boxit{0.3in} 57.0& 74.7& 79.8& 64.6& 74.1& 74.6& 65.2& 55.1& 81.0& 74.6& 59.7& 84.3& $\cellcolor[gray]{0.9}70.4$ \\

        \cellcolor[gray]{0.9}SPL \cite{spl}& & 54.5& 77.8& 81.9& 65.1& 78.0& 81.1& 66.0& 53.1& 82.8& 69.9& 55.3& \boxit{0.3in}86.0& $\cellcolor[gray]{0.9}71.0$ \\

        \cellcolor[gray]{0.9}SRDC \cite{srdc}& & 52.3& 76.3& 81.0& 69.5& 76.2& 78.0& 68.7& 53.8& 81.7& 76.3& \boxit{0.3in}57.1& 85.0& $\cellcolor[gray]{0.9}71.3$ \\

        \cellcolor[gray]{0.9}CLIP \cite{clip}& & 51.6& 81.9& 82.6& 71.9& 81.9& 82.6& 71.9& 51.6& 82.6& 71.9& 51.6& 81.9& $\cellcolor[gray]{0.9}72.0$ \\

         \cellcolor[gray]{0.9}DAPL \cite{dapl}& & 54.1& 84.3& 84.8& 74.4& 83.7& 85.0& 74.5& 54.6& 84.8& 75.2& 54.7& 83.8& $\cellcolor[gray]{0.9}74.5$ \\


         \cellcolor{blue!15}\textsc{AD-CLIP}& & \multirowcell{-8}[-5.0ex]{\hspace*{-0.8em} \vspace{2.8em}\turnbox{90}{\thead{RN-50}}} \hspace{-1.0cm} 55.4& \boxit{0.3in}85.2& \boxit{0.3in}85.6& \boxit{0.3in}76.1& \boxit{0.3in}85.8& \boxit{0.3in}86.2& \boxit{0.3in}76.7& \boxit{0.3in}56.1& \boxit{0.3in}85.4& \boxit{0.3in}76.8& 56.1& 85.5& $\cellcolor{blue!15}\textbf{75.9} \pm 0.1$ \\
\midrule    
        \cellcolor[gray]{0.9}CDTrans* \cite{cdtrans}& & 68.8& 85.0& 86.9& 81.5& 87.1& 87.3& 79.6& 63.3& 88.2& 82.0& 66.0& 90.6& $\cellcolor[gray]{0.9}80.5$\\

        \cellcolor[gray]{0.9}TVT \cite{tvt}& & 74.9& 86.8& 89.5& 82.8& 88.0& 88.3& 79.8& 71.9& 90.1& 85.5& 74.6& 90.6& $\cellcolor[gray]{0.9}83.6$\\

        \cellcolor[gray]{0.9}SSRT \cite{ssrt}& & \boxit{0.3in}75.2& 89.0& 91.1& 85.1& 88.3& 90.0& 85.0& 74.2& 91.3& 85.7& \boxit{0.3in}78.6& 91.-8& $\cellcolor[gray]{0.9}85.4$\\

        \cellcolor[gray]{0.9}CLIP \cite{clip} & & 67.8& 89.0& 89.8 & 82.9& 89.0& 89.8& 82.9& 67.8& 89.8& 82.9& 67.8& 89.0& $\cellcolor[gray]{0.9}82.4$  \\

        \cellcolor[gray]{0.9}DAPL \cite{dapl} & & 70.6& 90.2& 91.0& 84.9& 89.2& 90.9& 84.8& 70.5& 90.6& 84.8& 70.1& 90.8& $\cellcolor[gray]{0.9}84.0$ \\ 

        \cellcolor{blue!15}\textsc{AD-CLIP}& & \multirowcell{-6}[-4.8ex]{\hspace*{-1.8em} \vspace{1.5em}\turnbox{90}{\thead{ViT-B/16}}} \hspace{-1.0cm} 70.9& \boxit{0.3in}92.5& \boxit{0.3in}92.1& \boxit{0.3in}85.4& \boxit{0.3in}92.4& \boxit{0.3in}92.5& \boxit{0.3in}86.7& \boxit{0.3in}74.3& \boxit{0.3in}93.0& \boxit{0.3in}86.9& 72.6& \boxit{0.3in}93.8& $\cellcolor{blue!15}\textbf{86.1} \pm 0.2$ \\ 
        
    \midrule
        \cellcolor[gray]{0.9}CLIP \cite{clip} & & 74.2& 93.1& 93.3& 87.3& 93.1& 93.3& 87.3& 74.2& 93.3& 87.3& 74.2& 93.1& $\cellcolor[gray]{0.9}87.0$  \\


        \cellcolor[gray]{0.9}DAPL \cite{dapl}& & 77.3& 94.6& 94.3& 88.6& 94.6& 94.0& 88.8& 76.8& 94.0& 89.0& 77.8& 94.4& $\cellcolor[gray]{0.9}88.7$ \\

        \vspace{2mm}


         
        \hspace{-0.1cm}\cellcolor{blue!15}\textsc{AD-CLIP} & & \multirowcell{-3}[-0.0ex]{\hspace*{-1.5em} \vspace{-3.7em}\turnbox{90}{\thead{ViT-L/14}}} \hspace{-1.0cm} \textbf{80.3}& \textbf{95.4} & \textbf{95.7} & \textbf{90.9} & \textbf{95.5} & \textbf{95.2} & \textbf{90.1} & \textbf{79.6} & \textbf{95.1} & \textbf{90.8} & \textbf{81.1} & \textbf{95.9} & $\cellcolor{blue!15}\textbf{90.5} \pm0.2$ \\\hline
-
        
        
    \end{tabular}
    }
    \label{tab:officehome}
\end{table*}



However, we have observed that on the VisDA-2017 dataset, \textsc{AD-CLIP} was not able to outperform the best results from different models on 7 out of 12 classes when ResNet-101 was used as the vision encoder backbone. In particular, traditional Transformer-based methods (SSRT and CDTrans) achieved the overall best results on 4 classes, namely \textit{car}, \textit{knife}, \textit{person}, and \textit{plant}. Nevertheless, the average performance of \textsc{AD-CLIP} still outperformed DAPL by $0.8\%$, $1.2\%$, and $2.2\%$ when using ResNet-101, ViT-B/16, and ViT-L/14 as backbones, respectively. On the Mini-DomainNet dataset, we utilized CLIP and DAPL as baselines for comparison with \textsc{AD-CLIP}. The results presented in Table \ref{tab:minidomainnet} illustrate that \textsc{AD-CLIP} outperformed both baseline methods by a considerable margin across all backbone models. The comprehensive evaluation indicates that \textsc{AD-CLIP} is a competitive method for domain adaptation tasks, achieving notable performance improvements on multiple benchmark datasets. While its performance on some classes of the VisDA-2017 dataset exhibited slight limitations, \textsc{AD-CLIP} remains a powerful approach, consistently outperforming existing methods on diverse datasets and backbone configurations. The results on Mini-DomainNet further validate \textsc{AD-CLIP}'s efficacy as a reliable choice for addressing domain shift challenges. Overall, the outcomes underscore the potential of \textsc{AD-CLIP} as an effective and versatile tool in the domain adaptation landscape.

\begin{table*}[!ht]
    \centering
    \caption{\textcolor{royalblue}{Comparison of \textsc{AD-CLIP} with state-of-the-art methods for UDA task on VisDA-2017 \cite{visda} dataset. We show our results for every class with three different vision backbones. However, CDTrans* has used DeiT-base \cite{deit} backbone only. The overall best accuracy and best within per backbone are indicated in bold and box respectively.}}\label{tab:visda}
    \vspace{0.2cm}
    \scalebox{0.6}{
    \begin{tabular}{lccccccccccccccc}
    \toprule
        
       \rowcolor{gray!20} {\textbf{Method}} & $f_v$ & {plane} & {bicycle} & {bus} & {car} & {horse} & {knife} & {mcycl} & {person} & {plant} & {sktbrd} & {train} & {truck} & {Avg} \\ 
        \midrule

        \cellcolor[gray]{0.9}RN-101 \cite{resnet} & & 55.1& 53.3& 61.9& 59.1& 80.6& 17.9& 79.7& 31.2& 81.0& 26.5& 73.5& 8.5& \cellcolor[gray]{0.9}52.4 \\

        \cellcolor[gray]{0.9}DANN \cite{dann} & & 81.9& 77.7& 82.8& 44.3& 81.2& 29.5& 65.1& 28.6& 51.9& 54.6& 82.8& 7.8& \cellcolor[gray]{0.9}57.4 \\

        \cellcolor[gray]{0.9}JAN \cite{jan} & & 75.7& 18.7& 82.3& 86.3& 70.2& 56.9& 80.5& 53.8& 92.5& 32.2& 84.5& 54.5& \cellcolor[gray]{0.9}65.7 \\ 

        \cellcolor[gray]{0.9}MODEL \cite{model} & & 94.8& 73.4& 68.8& 74.8& 93.1& \boxit{0.3in}95.4& 88.6& \boxit{0.3in}84.7& 89.1& 84.7& 83.5& 48.1& \cellcolor[gray]{0.9}81.6 \\ 

        \cellcolor[gray]{0.9}STAR \cite{star} & & 95.0& \boxit{0.3in}84.0& 84.6& 73.0& 91.6& 91.8& 85.9& 78.4& \boxit{0.3in}94.4& 84.7& 87.0& 42.2& \cellcolor[gray]{0.9}82.7 \\ 

        \cellcolor[gray]{0.9}CLIP \cite{clip} & & \boxit{0.3in}98.2& 83.9& 90.5& 73.5& 97.2& 84.0& 95.3& 65.7& 79.4& 89.9& 91.8& 63.3& \cellcolor[gray]{0.9}84.4  \\

        \cellcolor[gray]{0.9}DAPL \cite{dapl} & & 97.8& 83.1& 88.8& \boxit{0.3in}77.9& 97.4& 91.5& 94.2& 79.7& 88.6& 89.3& \boxit{0.3in}92.5& 62.0&  \cellcolor[gray]{0.9}86.9  \\

        \cellcolor{blue!15}\textsc{AD-CLIP} & & \multirowcell{-8}[-5.0ex]{\hspace*{0.3em} \vspace{2.4em}\turnbox{90}{\thead{RN-101}}} \hspace{-0.5cm} 98.1& 83.6& \boxit{0.3in}91.2& 76.6& \boxit{0.3in}98.1& 93.4& \boxit{0.3in}96.0& 81.4& 86.4& \boxit{0.3in}91.5& 92.1& \boxit{0.3in}64.2& $\cellcolor{blue!15}\textbf{87.7} \pm 0.2$ \\
        \midrule

        \cellcolor[gray]{0.9}CDTrans* \cite{cdtrans}& & 97.1& 90.5& 82.4& 77.5& 96.6& 96.1& 93.6& \textbf{88.6}& \textbf{97.9}& 86.9& 90.3& 62.8& \cellcolor[gray]{0.9}88.4 \\ 

        \cellcolor[gray]{0.9}TVT \cite{tvt}& & 97.1& \boxit{0.3in}92.9& 85.3& 66.4& 97.1& 97.1& 89.3& 75.5& 95.0& 94.7& 94.5& 55.1& \cellcolor[gray]{0.9}86.7\\

        \cellcolor[gray]{0.9}SSRT \cite{ssrt}& & 98.9& 87.6& 89.1& \textbf{84.8}& 98.3& \textbf{98.7}& 96.3& 81.1& 94.9& \boxit{0.3in}97.9& 94.5& 43.1& \cellcolor[gray]{0.9}88.8\\

        \cellcolor[gray]{0.9}CLIP \cite{clip} & & 99.1& 91.7& 93.8 & 76.7& 98.4& 91.7& 95.3& 82.7& 86.5& 96.0& 94.6& 60.5& \cellcolor[gray]{0.9}88.9  \\

        \cellcolor[gray]{0.9}DAPL \cite{dapl} & & 99.2& 92.5& 93.3& 75.4& 98.6& 92.8& 95.2& 82.5& 89.3& 96.5& 95.1& 63.5& \cellcolor[gray]{0.9}89.5 \\ 

        \hspace{-0.01cm}\cellcolor{blue!15}\textsc{AD-CLIP} & & \multirowcell{-6}[-4.8ex]{\hspace*{-1.0em} \vspace{1.5em}\turnbox{90}{\thead{ViT-B/16}}} \hspace{-0.5cm} \boxit{0.3in}99.6& 92.8& \boxit{0.3in}94.0& 78.6& \boxit{0.3in}98.8& 95.4& \boxit{0.3in}96.8& 83.9& 91.5& 95.8& \boxit{0.3in}95.5& \boxit{0.3in}65.7& $\cellcolor{blue!15}\textbf{90.7} \pm 0.3$ \\
        \midrule
 
        \cellcolor[gray]{0.9}CLIP \cite{clip} & & 99.5& 91.1& 92.0& 69.2& 99.2& 89.5& 97.5& 84.3& 82.8& 98.2& 96.9& 69.1& \cellcolor[gray]{0.9}89.1  \\

        \cellcolor[gray]{0.9}DAPL \cite{dapl} & & 99.6& 91.6& 92.9& 75.7& 99.4& 93.3& 97.4& 84.8& 85.5& 97.9& 97.4& 70.5& \cellcolor[gray]{0.9}90.5 \\ 


        \vspace{2mm}

        \hspace{-0.12cm}\cellcolor{blue!15}\textsc{AD-CLIP} & & \multirowcell{-3}[-0.0ex]{\hspace*{-0.6em} \vspace{-3.7em}\turnbox{90}{\thead{ViT-L/14}}} \hspace{-0.5cm} \textbf{99.8}& \textbf{93.2} & \textbf{95.2} & \boxit{0.3in}79.1 & \textbf{99.7} & \boxit{0.3in}96.4 & \textbf{98.5} & \boxit{0.3in}86.4 & \boxit{0.3in}94.0 & \textbf{98.6} & \textbf{98.1} & \textbf{73.2} & $\cellcolor{blue!15}\textbf{92.7} \pm 0.1$ \\\bottomrule
    \end{tabular}}
\end{table*}

\subsection{Ablation analysis}
\begin{figure}[h]
    \centering
    \includegraphics[scale=0.3]{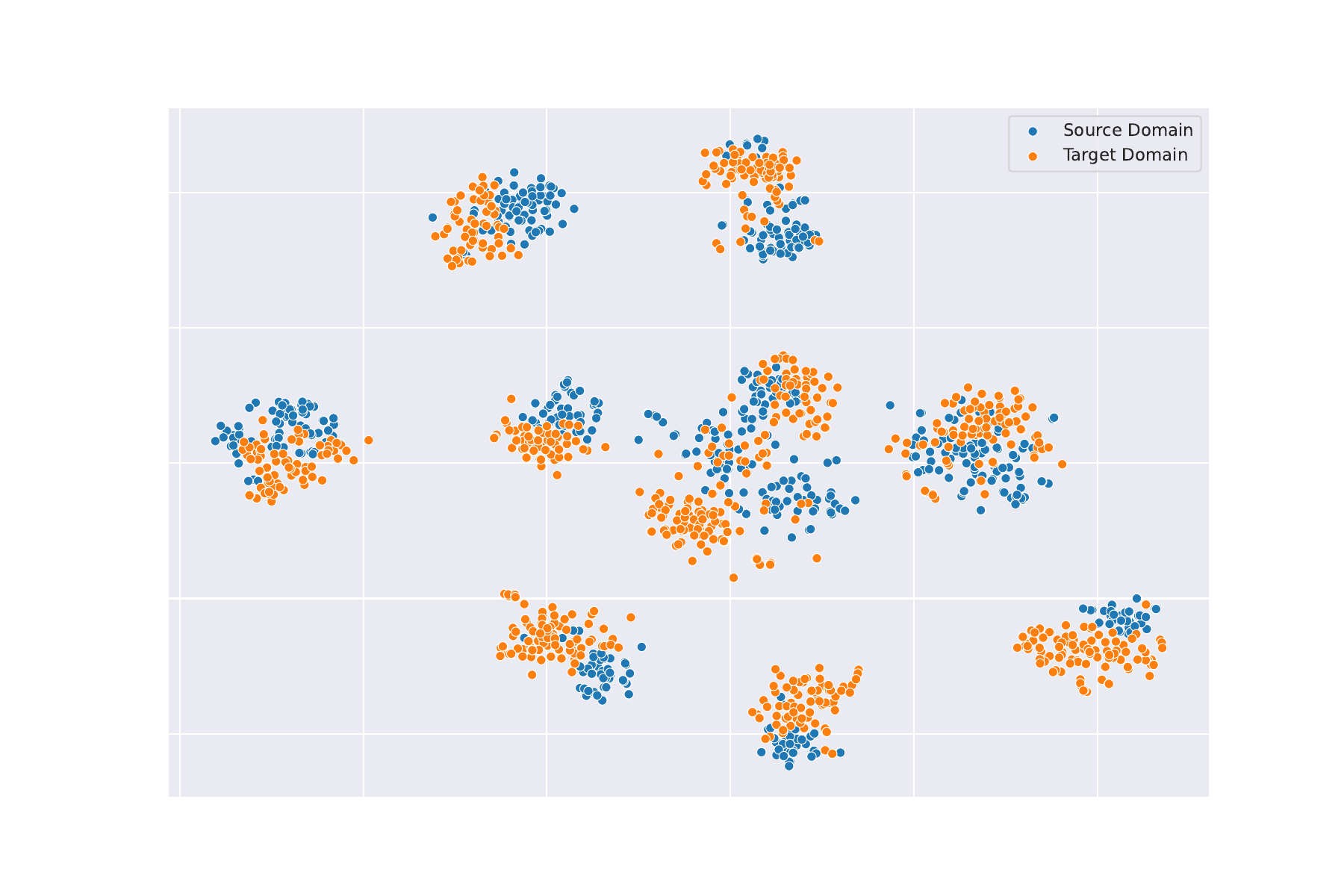}
    \vspace{0.1cm}
    \caption{\textcolor{royalblue}{t-SNE visualizations of text embeddings from art and clipart domains of 10 classes of Office-Home.}}
    \label{fig:tsne}
\end{figure}

\noindent \textbf{t-SNE \cite{tsne} visualization}: In this section, we conduct an ablation study to gain insights into the domain invariance and discriminativeness achieved by our proposed model, \textsc{AD-CLIP}. Specifically, we visualize the t-SNE representations of the text embeddings corresponding to the art and clipart domains across the 10 classes of the Office-Home dataset \cite{officehome}. Figure \ref{fig:tsne} exhibits the t-SNE visualization of the text embeddings generated by \textsc{AD-CLIP} for the art and clipart domains. The visualization provides an intuitive representation of the distribution and clustering of textual features across different classes and domains.

The t-SNE visualization highlights the simultaneous achievement of domain invariance and discriminativeness by \textsc{AD-CLIP}. The textual embeddings show clear separation between different classes, indicating the discriminative power of the model in distinguishing diverse object categories. Moreover, despite the domain shift between the art and clipart domains, the embeddings demonstrate overlapping regions, signifying the successful domain invariance achieved by \textsc{AD-CLIP}. This capability to maintain similarity between text embeddings from different domains is crucial for effective domain adaptation.
\vspace{0.3cm}

\begin{figure}[h]
    \centering
    \includegraphics[scale=0.45]{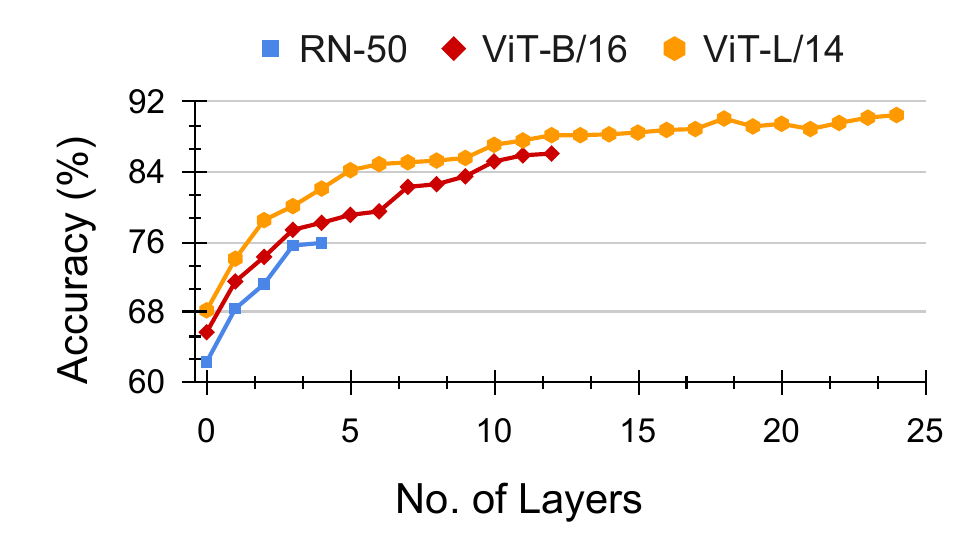}
    \vspace{0.1cm}
    \caption{\textcolor{royalblue}{Performance of \textsc{AD-CLIP} with different layers of RN50, ViT-B/16 and ViT-L/14 backbones to extract multi-scale features on Office-Home.}}
    \label{fig:multi}
\end{figure}

\noindent \textbf{Sensitivity to the multi-scale features:} We also conduct an ablation study to evaluate the sensitivity of \textsc{AD-CLIP} to visual content and style features obtained from different layers of the CLIP vision backbones. Our aim is to investigate the impact of incorporating multiple $f_v$ layers on the performance of \textsc{AD-CLIP}. Specifically, we utilize three CLIP vision backbones, namely ResNet-50, ViT-B/16, and ViT-L/14, and vary the number of feature layers used to calculate $\hat{\mathcal{F}}(x)$, ranging from the initial to the final layer. To perform the ablation study, we modify the feature extraction process in \textsc{AD-CLIP} by leveraging different $f_v$ layers from the chosen CLIP vision backbones. We begin by extracting visual features from the initial layer and progressively increase the number of layers until reaching the final layer. We then evaluate the impact of these variations on the overall performance of \textsc{AD-CLIP}. Figure \ref{fig:multi} presents the results of our ablation study. The plot illustrates the performance trend of \textsc{AD-CLIP} as we incorporate additional layers to extract content features. Remarkably, the results demonstrate a consistent upward trend, indicating that including more layers for content feature extraction leads to improved performance. This observation suggests that the model benefits from incorporating visual information from multiple layers, enabling it to capture more nuanced and discriminative features.

\begin{table*}[!ht]

    \centering
    \caption{\textcolor{royalblue}{Comparison of \textsc{AD-CLIP} with the state-of-the-art vision-language models for UDA task on Mini-DomainNet \cite{domain_net} dataset. The overall best accuracy and best within per backbone are indicated in bold and box respectively.}}
    \vspace{0.3cm}
    \scalebox{0.6}{
    \begin{tabular}{lccccccccccccccc}
    \toprule
        
       \rowcolor{gray!20} {\textbf{Method}} & $f_v$ & {Cl$\rightarrow$Pn} & {Cl$\rightarrow$Rl} & {Cl$\rightarrow$Sk} & {Pn$\rightarrow$Cl} & {Pn$\rightarrow$Rl} & {Pn$\rightarrow$Sk} & {Rl$\rightarrow$Cl} & {Rl$\rightarrow$Pn} & {Rl$\rightarrow$Sk} & {Sk$\rightarrow$Cl} & {Sk$\rightarrow$Pn} & {Sk$\rightarrow$Rl} & {\textbf{Avg}} \\ 
        \midrule

        \cellcolor[gray]{0.9}CLIP \cite{clip}& & 67.9& 84.8& 62.9& 69.1& 84.8& 62.9& 69.2& 67.9& 62.9& 69.1& 67.9& 84.8& $\cellcolor[gray]{0.9}71.2$ \\

         \cellcolor[gray]{0.9}DAPL \cite{dapl}& & \boxit{0.3in}72.4& 87.6& 65.9& 72.7& \boxit{0.3in}87.6& \boxit{0.3in}65.6& 73.2& 72.4& 66.2& \boxit{0.3in}73.8& 72.9& 87.8& $\cellcolor[gray]{0.9}74.8$ \\
         \vspace{2mm}
         \hspace{-0.2cm} \cellcolor{blue!15}\textsc{AD-CLIP}& & \multirowcell{-3}[-0.0ex]{\hspace*{-1.6em} \vspace{-2.5em}\turnbox{90}{\thead{RN-50}}} \hspace{-1.0cm} 71.7& \boxit{0.3in}88.1& \boxit{0.3in}66.0& \boxit{0.3in}73.2& 86.9& 65.2& \boxit{0.3in}73.6& \boxit{0.3in}73.0& \boxit{0.3in}68.4& 72.3& \boxit{0.3in}74.2& \boxit{0.3in}89.3& $\cellcolor{blue!15}\textbf{75.2} \pm 0.2$ \\ 
\midrule    

        \cellcolor[gray]{0.9}CLIP \cite{clip} & & 80.3& 90.5& 77.8& 82.7& 90.5& 77.8& 82.7& 80.3& 77.8& 82.7& 80.3& 90.5& $\cellcolor[gray]{0.9}82.8$  \\

        \cellcolor[gray]{0.9}DAPL \cite{dapl} & & 83.3& 92.4& 81.1& 86.4& 92.1& 81.0& 86.7& 83.3& 80.8& 86.8& 83.5& 91.9& $\cellcolor[gray]{0.9}85.8$ \\

        \vspace{2mm}

        \hspace{-0.2cm} \cellcolor{blue!15}\textsc{AD-CLIP}& & \multirowcell{-3}[-0.0ex]{\hspace*{-1.6em} \vspace{-3.7em}\turnbox{90}{\thead{ViT-B/16}}} \hspace{-1.0cm} \boxit{0.3in}84.3& \boxit{0.3in}93.7& \boxit{0.3in}82.4& \boxit{0.3in}87.5& \boxit{0.3in}93.5& \boxit{0.3in}82.4& \boxit{0.3in}87.3& \boxit{0.3in}84.5& \boxit{0.3in}81.6& \boxit{0.3in}87.9& \boxit{0.3in}84.8& \boxit{0.3in}93.0& $\cellcolor{blue!15}\textbf{86.9} \pm 0.2$ \\   

    \midrule

        \cellcolor[gray]{0.9}CLIP \cite{clip} & & 85.2& 92.4& 86.2& 89.2& 92.4& 86.2& 89.2& 85.2& 86.2& 89.2& 85.2& 92.4& $\cellcolor[gray]{0.9}88.3$  \\
        \cellcolor[gray]{0.9}DAPL \cite{dapl}& & 86.8& 93.5& 87.9& 90.5& 93.5& 88.3& 90.2& 87.8& 88.6& 90.0& 86.8& 93.5& $\cellcolor[gray]{0.9}89.8$ \\
        \vspace{2mm}
        \hspace{-0.2cm} \cellcolor{blue!15}\textsc{AD-CLIP} & & \multirowcell{-3}[-0.0ex]{\hspace*{-1.6em} \vspace{-3.7em}\turnbox{90}{\thead{ViT-L/14}}} \hspace{-1.0cm} \textbf{89.1}& \textbf{94.5} & \textbf{89.2} & \textbf{91.9} & \textbf{95.0} & \textbf{90.1} & \textbf{92.0} & \textbf{89.2} & \textbf{90.3} & \textbf{92.3} & \textbf{88.4} & \textbf{95.1} & $\cellcolor{blue!15}\textbf{91.4} \pm 0.1$ \\     
        
\bottomrule
        
    \end{tabular}}
    \label{tab:minidomainnet}
\end{table*}

\begin{figure}[h]
    \centering
    \includegraphics[scale=0.4]{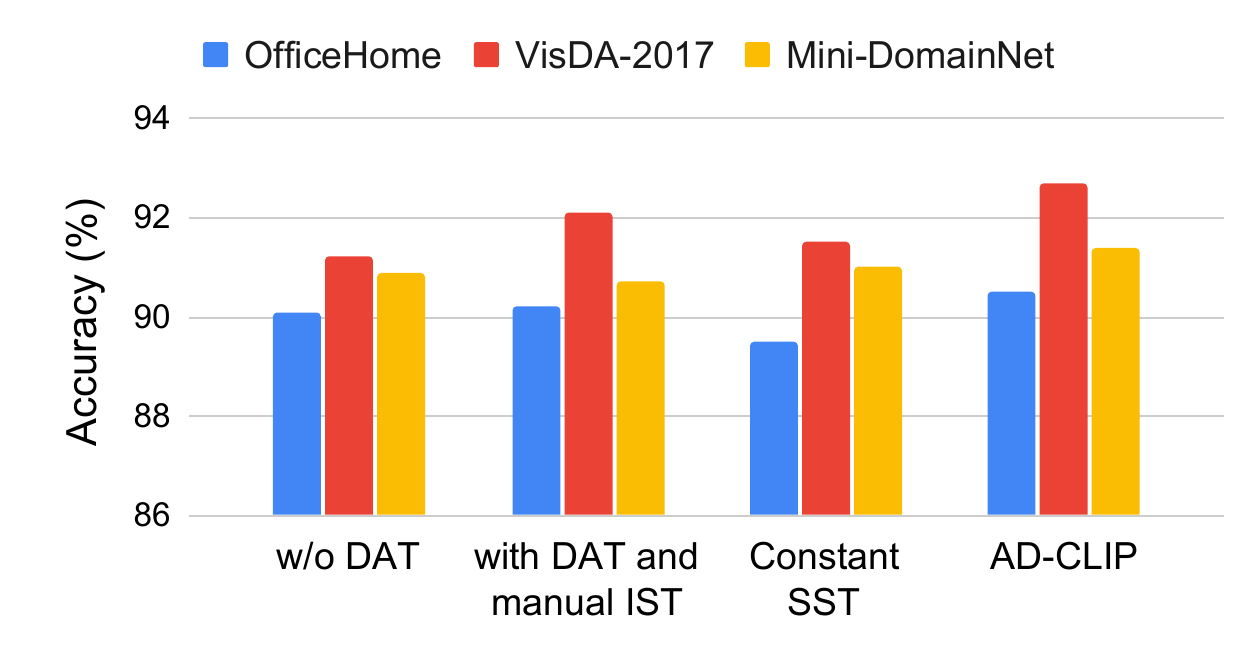}
    \caption{\textcolor{royalblue}{Comparison of results of \textsc{AD-CLIP} with different prompt settings. Here \textsc{DAT}, \textsc{IST} and \textsc{SST} refer to domain-agnostic token, image-specific tokens and source-domain style tokens.}}
    \label{fig:prompt}
\end{figure}

\noindent \textbf{Sensitivity on prompt behaviour and multi-scale feature information:} We have analyzed the impact of different prompt settings on the performance of \textsc{AD-CLIP}. The evaluation involved comparing various configurations, including the presence or absence of the domain agnostic token (DAT), the use of manual prompts with image-specific tokens (IST), the source-domain style token (SST) approach, and the full \textsc{AD-CLIP} model with the $f_{smn}$ mechanism. Figure \ref{fig:prompt} illustrates the results of the ablation study. We observed that omitting the DAT from the prompt led to a decrease in performance, highlighting its importance in guiding \textsc{AD-CLIP} to learn domain-invariant representations. When using manual prompts with IST instead of learned ones, minor improvements were observed, but the overall impact was not significant. Attempting to use an average style information from all source domains as the SST resulted in decreased performance and overfitting of the model. However, our full \textsc{AD-CLIP} model with $f_{smn}$ consistently achieved the best performance across all prompt settings, demonstrating its effectiveness in enhancing domain adaptation and discriminative representation learning. The ablation study provides valuable insights into the prompt configurations of \textsc{AD-CLIP}. The findings underscore the importance of the DAT, dynamic learning of IST and style information through $f_{smn}$, and highlight the superiority of our proposed \textsc{AD-CLIP} model in achieving robust domain adaptation performance.

\noindent \textbf{Sensitivity on loss terms:} Table \ref{tab:loss} presents the results of our ablation study on \textsc{AD-CLIP}, focusing on the influence of various loss terms across all three datasets. The experiments involved omitting the entropy minimization term and the KL divergence loss, both of which led to a significant decrease in performance, emphasizing their crucial role in the optimization process. On the other hand, utilizing all the loss functions consistently boosted the performance of \textsc{AD-CLIP}. Furthermore, we  have conducted the experiments under two settings: one involving multi-scale features and the other considering only the features from the final layer of $f_v$ to define content and style information. However, no significant differences in performance were observed between these settings, suggesting that both configurations are equally effective for enhancing \textsc{AD-CLIP}'s performance in domain adaptation tasks. The results confirm the significance of the entropy minimization term and KL divergence loss in the loss function of \textsc{AD-CLIP}. The adoption of these loss functions, in conjunction with multi-scale or final layer features, contributes to the model's robustness and demonstrates its potential as a versatile and effective approach for domain adaptation across diverse datasets.



\begin{figure}[h]
    \centering
    \includegraphics[scale=0.43]{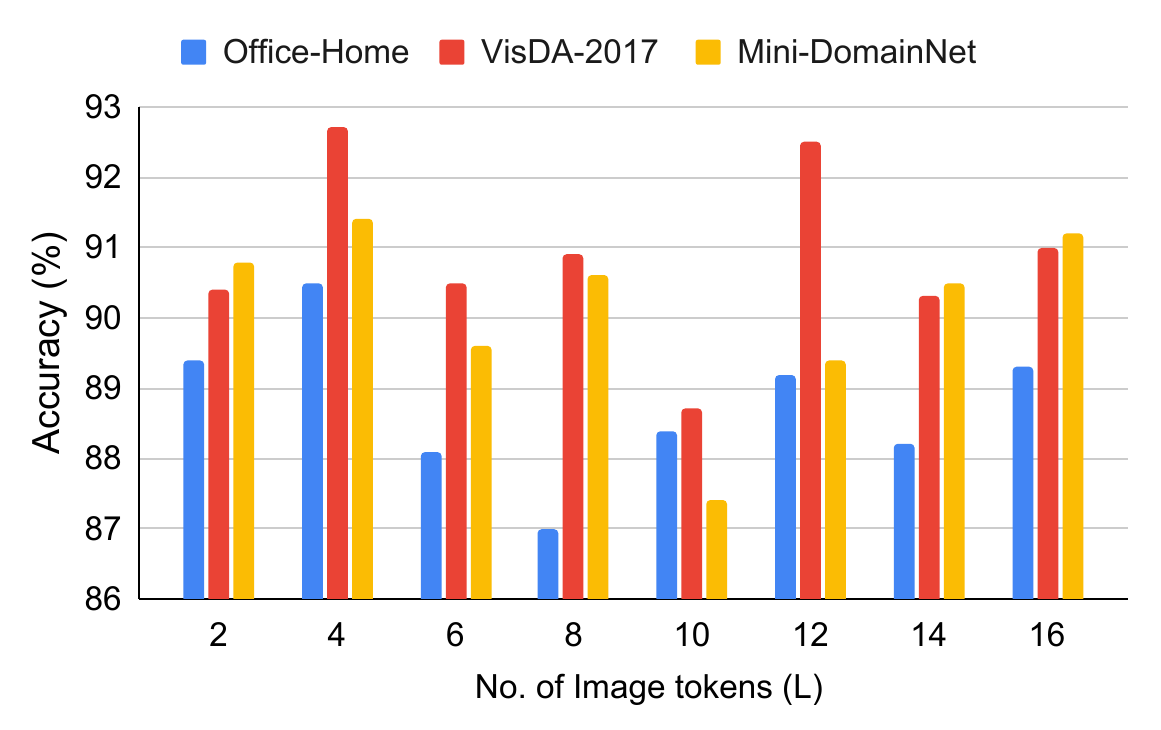}
    \caption{\textcolor{royalblue}{Performance of AD-CLIP with different numbers of image tokens in the prompt}}
    \label{fig:imageTokens}
\end{figure}

\vspace{0.2cm}
\noindent \textbf{Ablation analysis for Image token length:} In order to evaluate the sensitivity of AD-CLIP to the length of image-specific context tokens ${V_l}_{l=1}^L$ obtained from $C_v$, we conducted experiments on three benchmark datasets for unsupervised domain adaptation (UDA) tasks. Figure \ref{fig:imageTokens} displays the results of varying the length of image tokens.

The findings of the ablation study reveal that the best performance is achieved at an optimal length of $L$ = 4 for the image-specific context tokens. This indicates that AD-CLIP benefits from considering a moderate number of image tokens to effectively capture relevant visual information and enhance its performance in domain adaptation tasks across diverse datasets.

\begin{table}[!ht]
    \centering
    \caption{\textcolor{royalblue}{Ablation study of \textsc{AD-CLIP} with different losses in three datasets using source encoder ViT-L/14 and source-assisted encoder ViT-B/16. Here `w-ms' defines ablation with multi-scale features and `w/o-ms' defines without multi-scale features.}}
    \scalebox{0.67}{
    \begin{tabular}{lccccccc}
    \toprule
       \rowcolor{gray!20} & \multicolumn{2}{c}{Office-Home} & \multicolumn{2}{c}{VisDA-2017} & \multicolumn{2}{c}{Mini-DomainNet} \\ \cmidrule(lr){2-3} \cmidrule(lr){4-5}\cmidrule(lr){6-7}
        
       \rowcolor{gray!20} \multirow{-2}{*}{\textbf{Loss}} & {w-ms} & {w/o-ms} & {w-ms} & {w/o-ms} & {w-ms} & {w/o-ms} \\ 
        \midrule
         \cellcolor[gray]{0.9}$\mathbf{L}_{ce}$ (no adaptation) & 87.6& 87.4& 89.4& 89.3& 88.7& 88.6 \\ 
         \cellcolor[gray]{0.9}$\mathbf{L}_{ce}$ + $\mathbf{L}_{smn}$ (no adaptation) & 87.9& 87.2& 90.1& 89.5& 89.3& 89.2 \\
         \cellcolor[gray]{0.9}$\mathbf{L}_{ce}$ + $\mathbf{L}_{smn}$ + $\mathbf{L}_{em}$ & 88.1& 87.7& 89.8& 89.5& 89.6& 89.4 \\
         \cellcolor[gray]{0.9}$\mathbf{L}_{ce}$ + $\mathbf{L}_{Align}$ & 89.1& 89.2& 91.0& 90.9& 90.6& 90.8 \\        
         \cellcolor[gray]{0.9}$\mathbf{L}_{ce}$ + $\mathbf{L}_{smn}$ + $\mathbf{L}_{Align}$ & \textbf{90.5}& 90.1& \textbf{92.7}& 91.9& \textbf{91.4}& 91.1 \\ 
        
\bottomrule
        
    \end{tabular}}
    \label{tab:loss}
\end{table}

\begin{figure}[h]
    \centering
    \includegraphics[scale=0.4]{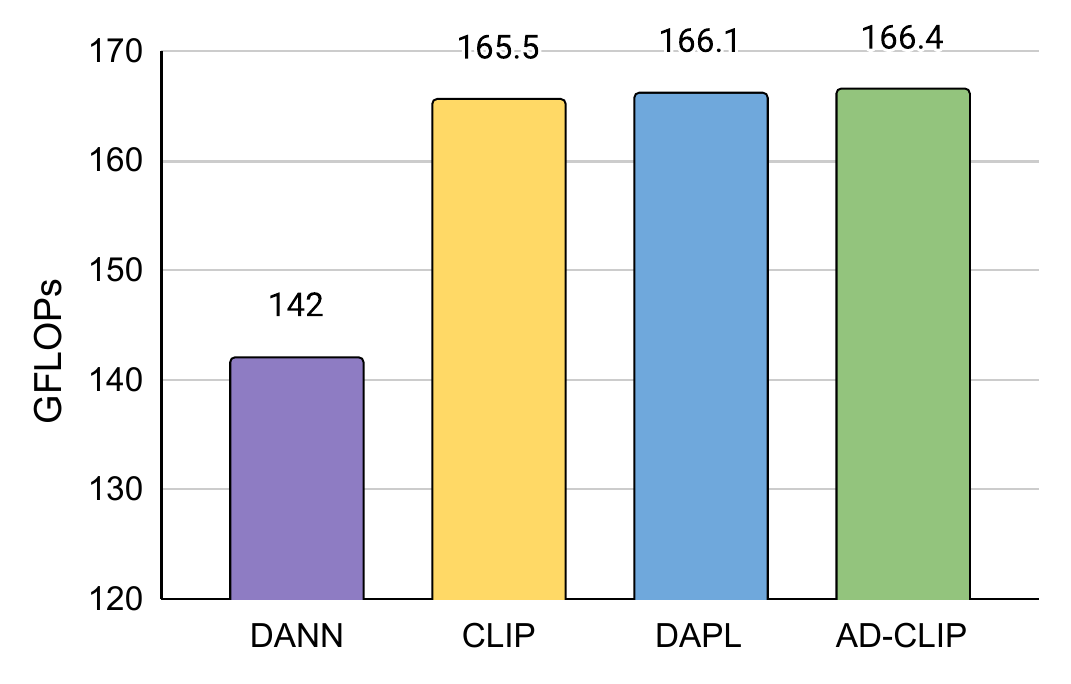}
    \caption{\textcolor{royalblue}{Comparison of the computational complexity of AD-CLIP with other UDA methods in terms of GFLOPs.}}
    \label{fig:gflops}
\end{figure}

\noindent\textbf{Model Complexity}
We train our model on NVIDIA RTX A6000 GPU with 48 GB card. Results indicate that AD-CLIP requires 17.2$\%$, 0.54$\%$, 0.18$\%$ more computational resources than DANN\cite{dann}, CLIP\cite{clip} and DAPL\cite{dapl} respectively as shown in Figure \ref{fig:gflops}. However, AD-CLIP outperforms every state-of-the-art method in UDA task.

\section{Takeaways}
In this paper, we propose a novel framework called \textsc{AD-CLIP} that tackles the unsupervised DA problem through prompt learning for foundation models. Our approach is based on the CLIP model and focuses on learning domain-invariant and class-generic prompt tokens using visual space features. To achieve this, we leverage the vision encoder of CLIP to extract multi-scale style and content features and adapt them to target datasets using learnable projector networks. Specifically, we learn three types of tokens in the prompts per image: domain token, image token, and class token. Additionally, we introduce a combination of distribution divergence loss and entropy minimization loss to align domains. Our experimental results on three benchmark DA datasets demonstrate that \textsc{AD-CLIP} outperforms existing state-of-the-art methods. In the future, we plan to extend our approach to solve specific applications such as person re-identification and medical imaging.
{\small
\bibliographystyle{ieee_fullname}
\bibliography{egbib}
}

\end{document}